\documentclass[10pt]{article}

\usepackage{graphicx}

\usepackage{csagh}
\usepackage{amsmath}
\usepackage{graphicx}
\usepackage{multirow}
\usepackage{booktabs}
\usepackage{caption}
\usepackage{subcaption}
\usepackage{xcolor}
\usepackage{pbox}

\begin{document}
\begin{opening}

\title{Melanoma Skin Cancer and Nevus Mole Classification using Intensity Value Estimation with Convolutional Neural Network}

\author[Department of Computer Science and Engineering, Dhaka University of Engineering \& Technology,  Gazipur-1707, Bangladesh, md.ashafuddula@gmail.com]{N. I. Md. Ashafuddula}

\author[Department of Computer Science and Engineering, Dhaka University of Engineering \& Technology,  Gazipur-1707, Bangladesh, rafiqul.islam@duet.ac.bd]{Rafiqul Islam}


\begin{abstract}
Melanoma skin cancer is one of the most dangerous and life-threatening cancer. Exposure to ultraviolet rays may damage the skin cell's DNA, which causes melanoma skin cancer. However, it is difficult to detect and classify melanoma and nevus mole at the immature stages. In this work, an automatic deep learning system is developed based on the intensity value estimation with a convolutional neural network model (CNN) to detect and classify melanoma and nevus mole more accurately. Since intensity levels are the most distinctive features for object or region of interest identification, the high-intensity pixel values are selected from the extracted lesion images. Incorporating those high-intensity features into the CNN improves the overall performance of the proposed model than the state-of-the-art methods for detecting melanoma skin cancer. To evaluate the system, we used 5-fold cross-validation. Experimental results show that a superior percentage of accuracy (92.58\%), sensitivity (93.76\%), specificity (91.56\%), and precision (90.68\%) are achieved.
\end{abstract}

\end{opening}

\section{Introduction}
\label{introLabel}
Skin covers the entire outside of our body and protects all of our internal body parts from environmental injuries. Nevertheless, because of its location, sometimes it suffers from different diseases that could often be life-threatening, and skin cancer is one of them \cite{55_N8_2_byrd2018human}. There are three major types of skin cancer: Basal Cell Carcinoma (BCC), Squamous Cell Carcinoma (SCC), and Melanoma and Merkel Cell Carcinoma (MCC) \cite{19_cancerorg}. Among all skin cancers, melanoma is the most dangerous and life-threatening one \cite{57_peram2019factorial}. It grows rapidly and can spread to other parts of the body. According to \cite{19_cancerorg}, about  99,780 new melanomas will be diagnosed (about 57,180 in men and 42,600 in women), and 7,650 people are expected to die of melanoma (about 5,080 men and 2,570 women) skin cancer in 2022. World Health Organization (WHO) reports one out of every three malignancies is affected by skin cancer \cite{50_N13_1_who}. From 2009  to 2019 there has been about a 55 percent increase in current skin cases examined annually \cite{8_vijayalakshmi2019melanoma}. 
The lifetime risk of being affected by melanoma is about 2.6\% (1 in 38) for whites, 0.1\% (1 in 1,000) for Blacks, and 0.6\% (1 in 167) for Hispanics \cite{19_cancerorg}. Early-stage detection of melanoma skin cancer can be beneficial in curing it \cite{42_eltayef2017detection,50_N13_5_silpa2013review,52_N4_dong2022deep,53_N4_al2018skin}. The early diagnosis survival rate of skin cancer is more than 90 percent \cite{50_N13_7_agilandeeswari2019skin}.

Due to the complex nature of skin lesions, dermatologists relate many visual observations, such as symmetry of the lesion area, size, shape, color, and border to diagnose malignant melanoma \cite{54_N5_adegun2019enhanced}. Sign of the Ugly Duckling lesion is another warning of Melanoma \cite{20_cancerorgabcde}. There are also some popular scoring techniques to identify malignant melanoma, where a 7-point checklist \cite{23_argenziano1998epiluminescence}, Menzies method, 3-point checklist, and ABCDE rules are the widely used method \cite{20_cancerorgabcde}.
Out of which a common way of detecting melanoma skin cancer is the ABCDE rule: A for asymmetry (one half of the mole does not match the other), B for border irregularity, C for Non-uniform color, D for diameter greater than 6 mm or ¼ inch (about the size of a pencil eraser), and E for evolving size, shape or color. To assess moles' size, color, and texture, specialist doctors examine lesion areas in which Total Dermatoscopic Score (TDS) is estimated using (A, B, C, D, E) and some weight factors. They conclude the diagnosis result from the TDS value. If the TDS is above a threshold value, then they identify the lesion as a malignant one otherwise benign \cite{7_mukherjee2019malignant}. 

Since this detection procedure is time-consuming and needs specialist doctors to conclude \cite{56_N10_kadampur2020skin}. There is a high possibility of being influenced by human subjectivity which makes it inconsistent in certain conditions. A computerized artificial intelligence-based system can easily detect and differentiate between melanoma and normal skin to eliminate this weakness. In recent times Deep Neural Networks (DNN) have been used in many different ways for medical imaging \cite{49_N22_haenssle2018man}. Convolutional Neural Network (CNN) is useful for image classification and recognition because of its capability to achieve high accuracy in a short time \cite{48_N2_zaman2021medical,51_N3_zaman2020analysis}. Compared to its predecessors, the main advantage of CNN is that it detects the important features automatically without any human supervision \cite{44_lakshminarayanan2022skin}. In this work, an intelligent system is developed to predict melanoma and nevus mole at an early stage, considering the complicated issue. Our innovative approach uses pre-processing steps, an intensity value estimation (IVE) model, and a CNN model. 

To train and test the model, we used five-fold cross-validation. The proposed model is evaluated using five well-known quantitative metrics: sensitivity, specificity, positive predictive value (PPV), negative predictive value (NPV), and accuracy. Considering the issues, the major contributions through this work can be enlisted as follows:
\begin{itemize}
    \item To develop an Intensity Value Estimation (IVE) model.
\item To present an efficient image resizing technique to keep the lesion shape, minimizing the possible data loss. 
 \item To design the proposed methodology by combining the resizing technique, IVE model, and Convolutional Neural Network (IVEwCNN) for automatic classification and detection of Melanoma skin cancer and Nevus mole.
\item A comparative histogram analysis is performed between our newly developed method and other state-of-the-art methods to evaluate the effectiveness of our newly developed model.
\end{itemize}

The rest of the paper is organized as follows: The related literature review is discussed in section \ref{litViewLabel}. The proposed methodology is categorized into subsections and briefly discussed in section \ref{ProMethodologyLabel}. Section \ref{highIntensePixelDetection} describes the high intense pixel value estimation and acquisition process. The experimental analysis is discussed in section \ref{expAnalysis}. Image acquisition and MED-NODE dataset descriptions are stated in section \ref{datasetLabel}. In section \ref{secExp} performance evaluation metrics and experimental results are discussed concerning different methods and the MED-NODE dataset. Finally, the conclusion is discussed in section \ref{secConc}.

\section{Background Study}
\label{litViewLabel}
Authors in \cite{4_giotis2015med} worked with the MED-NODE dataset. To segment regions of interest in healthy and lesioned areas, they used k-means clustering (k = 2). Before segmentation, a series of pre-processing steps are done to handle noise and illumination effectively. They utilized Gaussian smoothing ($\sigma$ = 5) and Kuwahara smoothing filters to remove noise and additional noisy features preserving edges. After removing noise, they mapped each image into 50 sub-image of pixel size (15 $\times$ 15). A cluster-based adaptive metric classifier was developed to extract 675 features per image. The training feature vector size was 2250 (50 $\times$ 45), and the evaluated feature vector size was 6250 (50 $\times$ 125). The model was good, but the dataset's size was one of its main limitations. 

A deep neural model in which the authors considered reducing illumination and noise effect in the pre-processing step was proposed in \cite{3_nasr2016melanoma}; then fed the enhanced images into a pre-trained CNN model. Hence, the dataset has limited images, so they used cropping and rotation to expand the training data. They produced a segmentation mask by applying the k-means classifier  (k = 2) to the pre-processed image. The mask was enhanced by applying morphological operations. Based on the information of the segmentation mask, a Gaussian filter was used on the standard skin parts. In their CNN model, 20 feature maps were generated in the first convolution layer, and 50 were generated in the second convolution layer. After each convolution layer, the pooling layer and a 2-layer fully connected stage were utilized. Finally, the diagnosis results were found from this 2-layer network with a linear transfer function. Training data is fed to the network using batch size 64. They randomly split the dataset into 20\% testing and 80\% training data with no overlapping. To train the model, they performed 20,000 iterations. They addressed that the Illumination correction increases their system's discrimination capability, which helped increase system accuracy. 

A combined model was developed by the authors in \cite{8_vijayalakshmi2019melanoma} using multi-level segmentation, CNN, support vector machine (SVM), and back-propagation neural network that combined Otsu, Modified Otsu, and watershed segmentation methods for segmentation. Finally, CNN and SVM were used for training and classification, respectively. Another authors in \cite{6_maiti2019improving} used MED-NODE and the international skin imaging collaboration (ISIC) dataset together with 2,170 images for their classification model. To avoid noise and common lighting problems, they used a few pre-processing steps to improve image texture and darken the perimeter of the lesion. They utilized Contrast enhancement and an anisotropic diffusion filter to correct the image contrast and remove noise and preserve lesion edges. Then, the shape features method and Principle Component Analysis (PCA) were used to extract features and reduce features dimension. Convolution Deep Neural Network (CDNN) was used where dropout is set 0.5 to avoid overfitting, learning rate tested 0.1 to 0.001, and discovered minor variation was working well, tested on different batch size but got 32 and 64 which works well in many situations. Their proposed system achieved 96.8\% accuracy with a few noticeable epochs in 0.41 min.

Another skin cancer detection technique was introduced in \cite{2_garg2019decision}  utilizing the well-known HAM-10000 dataset. In pre-processing steps, they removed noise, reduced image resolution, and applied image augmentation to avoid overfitting to increase the learnability of the system. The system produced several copies of existing images by various factors by applying translation, rotation, and zooming on images. For the classification task, they used CNN, the Transfer Learning method, with other classification algorithms such as XGBoost, SVM, and Random Forest Algorithms to classify and compare analysis. Finally, they yielded an accuracy of 90.51\% in the ResNet model's transfer learning approach. The author's used the MED-NODE dataset \cite{45_mukherjee2020malignant}. In the pre-processing step, Otsu's segmentation method is used on the grayscale image to segment the lesion part from the image. A total of 1900 features are extracted from each segmented lesion image. Twenty-five features are excluded from 1900 features as they are either too high or too low, or constant across the dataset. With the different types of training and cost functions available in a multi-layer neural network (MLP), this feature set is tested by various neurons. The relief method selects rank-wise best features among all 1,875 features. Using the features found from principal component analysis (PCA) of the extracted 1,875 features, Neural network (MLP), Linear SVM, Medium KNN, and Linear Discriminant are classifiers. Their work MLP with PCA feature only (used 25 features) shows 87.18\% accuracy. Authors \cite{24_Zheng2008} addressed the effect of contrast-enhancing, the image texture analysis, and considering pixel intensity values in an image classification model. They found that contrast-enhancing is designed to increase the discrimination between intensity values of an image. So, they can be easily identified by human and computer vision. The primary pieces of information are stored in the intensity value of a pixel. It can be a single value for a gray-level image or three values for a color image. They also addressed that image texture analysis could be an essential factor for pattern recognition due to the power of discrimination ability.

Noises were reduced from skin lesions using a Gaussian filter then applied an Improved K-means clustering-based segmentation \cite{review_a_corr_6_baldi2015}. The authors \cite{review_a_corr_6_baldi2015} extracted three different features using Local Binary Pattern (LBP), Grey Level Co-occurrence Matrix (GLCM), and RGB color channel features from the ROI of skin lesions. Using the extracted textural and color features from the lesion a distinctive hybrid super feature vector was created. For classification, they used a support vector machine (SVM), K-nearest neighbor (KNN), Naïve Bayes (NB), and Decision Tree (DT). The DERMIS dataset (146 are melanoma and 251 are nevus) was used in the experiment. The texture features, GLCM, and LBP features were merged with the color features to acquire a high classification accuracy.

\section{Proposed Methodology}
\label{ProMethodologyLabel}
\begin{figure}[!htb]
	\centering
		\includegraphics[width = 0.98\linewidth]{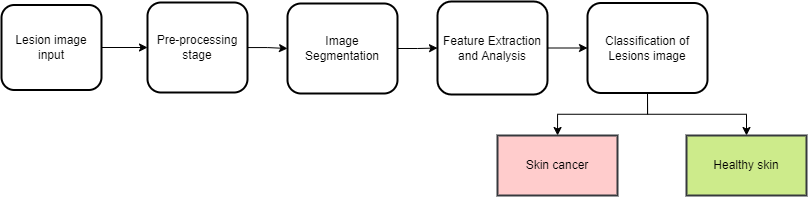}
	\centering
	    \caption{Process of diagnosing melanoma skin cancer}
	\label{FIG:GENERALDIAGRAM}
\end{figure}

The general process of detecting and diagnosing melanoma skin cancer has been summarized into key operating procedures such as image pre-processing, image segmentation, feature extraction, and analysis, and classification of lesions images \cite{54_N5_2_koundal2019advanced,54_N5_3_unver2019skin} as described in Fig. \ref{FIG:GENERALDIAGRAM}

\begin{figure}[!htb]
	\centering
		\includegraphics[width = 0.93\linewidth]{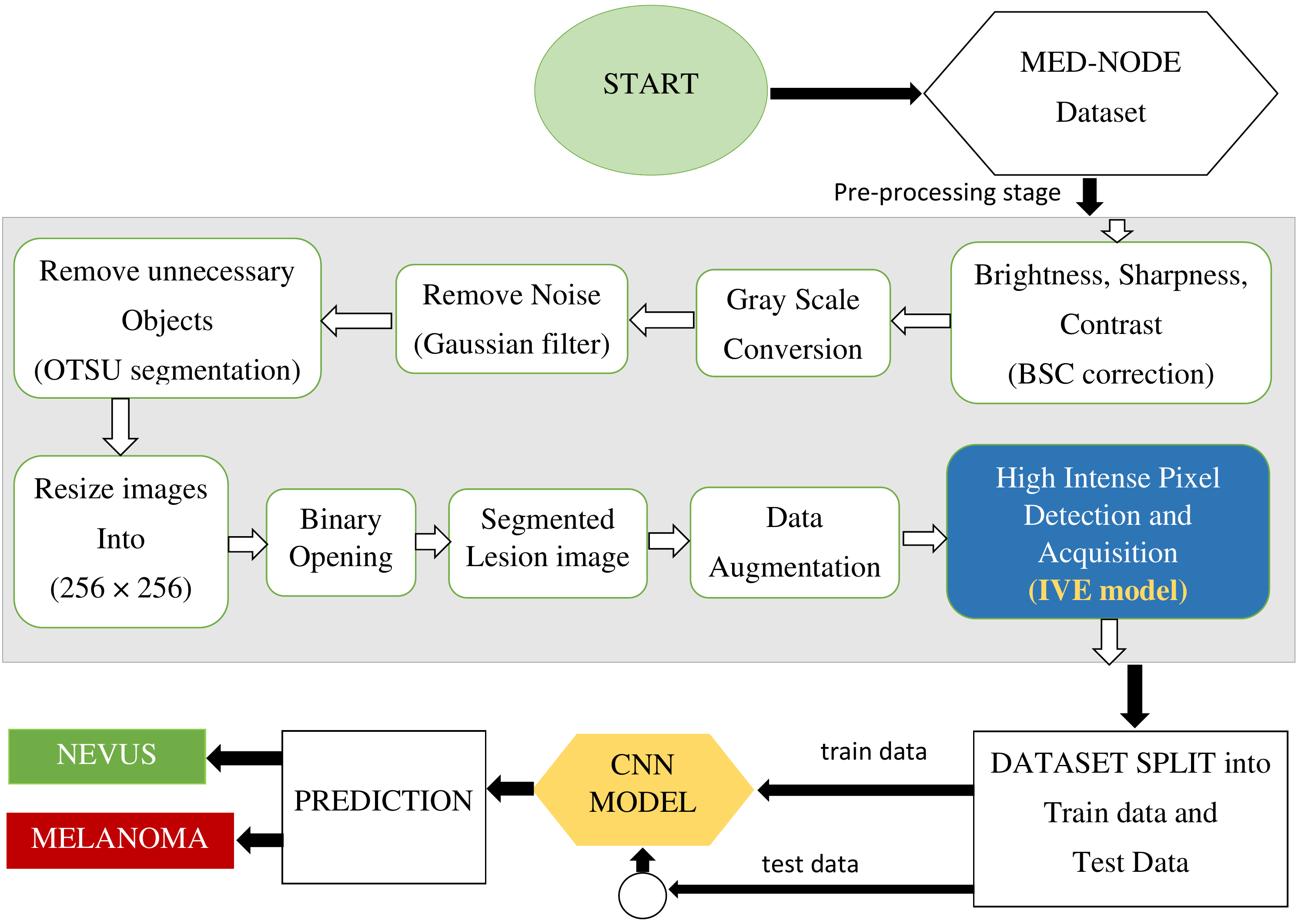}
	\centering
	    \caption{Detailed architecture of proposed methodology}
	\label{FIG:PROPOSEDMETHODLOGY}
\end{figure}

Our proposed methodology is divided into pre-processing, intensity value estimation, and convolutional neural network (CNN) steps. A few sub-parts in the pre-processing step to reduce artifacts that could mislead the CNN model. The proposed methodology is described by the block diagram shown in Fig. \ref{FIG:PROPOSEDMETHODLOGY}.

\subsection{Pre-processing}
\label{preprocessLabel}
In the pre-processing stage, images from the collected MED-NODE dataset \cite{4_giotis2015med} are passed through a series of basic image processing methods to reduce the effects of misleading factors on the CNN model. As images contain both the healthy skin and lesion parts, we corrected brightness (B), sharpness (S), and contrast (C) to enhance the images' lesion part to process further. In this experiment, a python ``ImageEnhance'' class is used from the ``pillow'' module to correct the BSC. The general intensity of the pixels refers to brightness in an image and the histogram of an image gives a clearer indication of the brightness. The higher end in the histogram indicates a brighter image and the image is darker when the histogram is confined to a small portion towards the lower end \cite{reviewer_corr_3_jaya2013iem}. The number of details present in an image refers to its sharpness. Lossy compression, motion blur, de-noising, and out-of-focus filtering are some of the causes that affect perceived image sharpness \cite{reviewer_corr_3_jaya2013iem}. In general, contrast refers to split-up the dark and bright regions in an image \cite{reviewer_corr_2_109_bora2017}. The contrast enhancement technique eliminates the anomaly that would otherwise occur between different regions in an image. 

An RGB image contains a three-channel red (R), green (G), and blue (B), whereas the grayscale image contains only one channel. The three-channel in an RGB image contains different intensity levels to represent that image in a color form. For various image processing tasks such as morphological operation segmentation, using a grayscale image is more accessible than an RGB image because it has only one channel. So, we converted the images from RGB to grayscale images. The grayscale pixel value is calculated as the weighted sum of the corresponding R (red), G (green), and B (blue) pixels as in the equation \ref{equ_rgb2gray} to convert an RGB channel image into a single grayscale channel image. Cathode-ray Tubes (CRT) phosphors use these weights to represent better the human perception of RGB than equal weights \cite{reviewer_corr_1_rgb2gray}. The texture in an image offers information about the spatial ordering of color or intensities in an image, or selected region of an image \cite{review_corr_4_yadav2014survey}. Since the grayscale image is formed from the three-channel RGB image using equation \ref{equ_rgb2gray}, the image color and texture could be represented by the different levels of pixel intensity in the grayscale image \cite{review_corr_5_kanan2012,review_corr_6_cadik2008}.

\begin{equation}
\label{equ_rgb2gray}
   G_{im} = 0.2125 R + 0.7154 G + 0.0721 B  
\end{equation}

Though high-resolution professional cameras take images, non-uniform lights create noise effects. We used the Gaussian filter to remove noise from grayscale lesion images ($\sigma$ = 1.35). 
The healthy skin part of the images is irrelevant to our model. So, before feeding the images into the CNN model, we segmented the lesion part from the healthy skin part using Otsu segmentation. 

\par\noindent\rule{\textwidth}{0.4pt}
\noindent\textbf{Algorithm 1: Resize Image}
\par\noindent\rule{\textwidth}{0.2pt}
\\
\textbf{Input: }\text{An 2D image $[oldIm_{r,c}]$, where $r \leq N_r$, $c \leq N_c$}\\
\textbf{Output: }\text{Resized 2D Image of size ($N_r \times N_c$)}
\begin{enumerate}
\label{algorithmresize}
    \item Initialization
    \begin{enumerate}
        \item $reqSize = (N_r, N_c)$
        \item An 2D image $[newIm_{r,c}]$, $r,c=reqSize$, where each element initialized with integer 0.
        \item $hr \leftarrow \lfloor (reqSize - len(oldIm_{row size}))/2 \rfloor$
        \item $hc \leftarrow \lfloor (reqSize - len(oldIm_{col size}))/2 \rfloor$
        \item $hcUp \leftarrow hc$
    \end{enumerate}
    \item \textbf{For{ $r \leftarrow 0$ to $len(oldIm_{row size})$} do}
    \begin{enumerate}
        \item \textbf{For{ $c \leftarrow 0$ to $len(oldIm_{col size})$} do}
        \begin{enumerate}
            \item $newIm[hr][hcUp] \leftarrow oldIm[r][c]$
            \item $hcUp \leftarrow hcUp +1$
        \end{enumerate}
        \item \textbf{End For}
        \item $hr \leftarrow hr +1$
        \item $hcUp \leftarrow hc$
    \end{enumerate}
    \item \textbf{End For}
    \item \textbf{Return $newIm$}
\end{enumerate}
\par\noindent\rule{\textwidth}{0.4pt}\bigskip

As the use of a conventional resizing algorithm on the grayscale image directly would distort the skin session shape \cite{47_salih2020skin}. So for convenience, keeping all the lesion shapes the same as the origin, we resized all The segmented masks (SM) and Gaussian filtered images (GFI) into a unique size ($N_r \times N_c$) (where $N_r = 256$ and $N_c = 256$) using Algorithm 1. 
We applied a binary opening on the resized mask images to remove hair, small objects, and unnecessary things to enhance the lesion area. The segmented mask is used to segment the lesion area (SLA) from the image following equation \ref{segment_musk_eqn} where ``r'' and ``c'' represents spatial (plane) coordinates of a 2D image, and the amplitude (r, c) is called the intensity or gray level at the point for that function.

\begin{equation}
\label{segment_musk_eqn}
   SLA_{r,c} = GFI_{r,c} * SM_{r,c} \quad \textrm{where}, (r,c) = 1, 2, \cdots, (N_c, N_r)  
\end{equation}

Data augmentation is a strategy that significantly increases the diversity and amount of data from available data to train a model without collecting new data. Cropping, Padding, Flipping, Zooming, and Varying the image size are the techniques that are used to augment the data. In this work, we used zoom in, zoom out, flipping, random contrast, random brightness, and rotation techniques to expand the data \cite{12_refianti2019classification,1_kassani2019comparative}. Before feeding the images into the proposed CNN model, the proposed IVE model is applied to take high intense pixel values from segmented lesion images. The model is divided into two sub-steps: edge detection and high intense pixel value estimation and acquisition. For edge detection, we used Canny Edge detection. The Canny Edge detection first eliminates the image noise by smoothing it and then finds the image gradient to highlight regions with high-spatial derivatives \cite{27_Jemal2010}. In the Canny Edge detection, to avoid losing precision while converting the image pixel values to unsigned 8-bit, we multiplied the segmented lesion image by 255. After data augmentation, the resized segmented lesion images are fed into the IVE model. The IVE model outputs a high intense pixel value stated in Fig. (\ref{sg}) and Fig. (\ref{sh}).

\subsection{Intensity Value Estimation}
\label{highIntensePixelDetection}
\begin{figure*}[htb]
     \centering
     \begin{subfigure}[b]{0.48\textwidth}
         \centering
          \includegraphics[width=.80\linewidth]{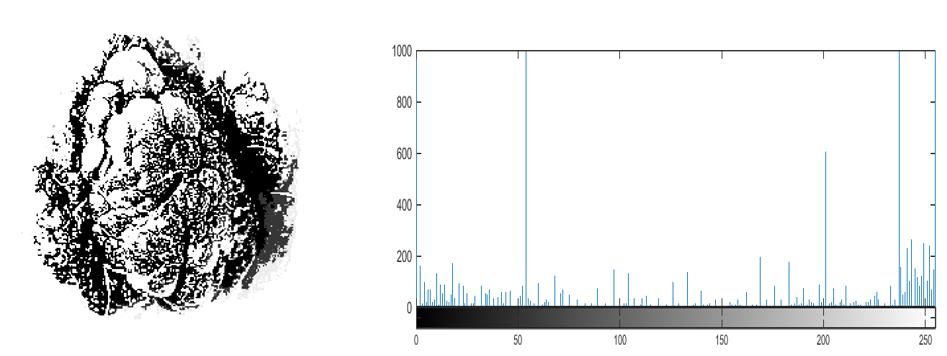}
         \caption{Gray level Image (Melanoma)}
         \label{sa}
     \end{subfigure}
     \hfill
     \begin{subfigure}[b]{0.48\textwidth}
         \centering
         \includegraphics[width=.80\linewidth]{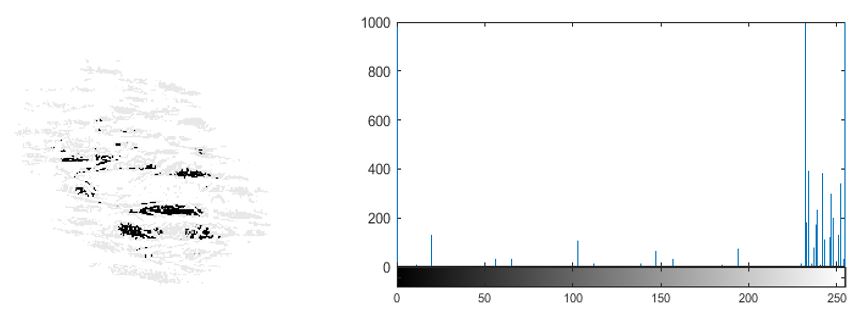}
         \caption{Gray level Image (Nevus)}
         \label{sb}
     \end{subfigure}
     \\
         \begin{subfigure}[b]{0.48\textwidth}
         \centering
          \includegraphics[width=.80\linewidth]{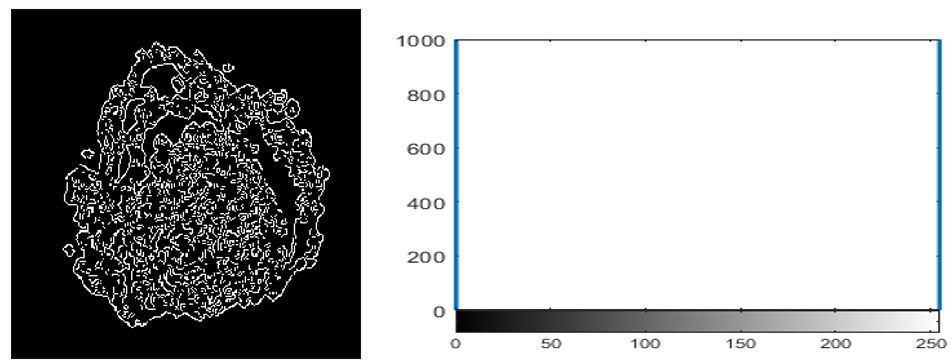}
         \caption{Canny Edge detection (Melanoma)}
         \label{sc}
     \end{subfigure}
     \hfill
     \begin{subfigure}[b]{0.48\textwidth}
         \centering
         \includegraphics[width=.80\linewidth]{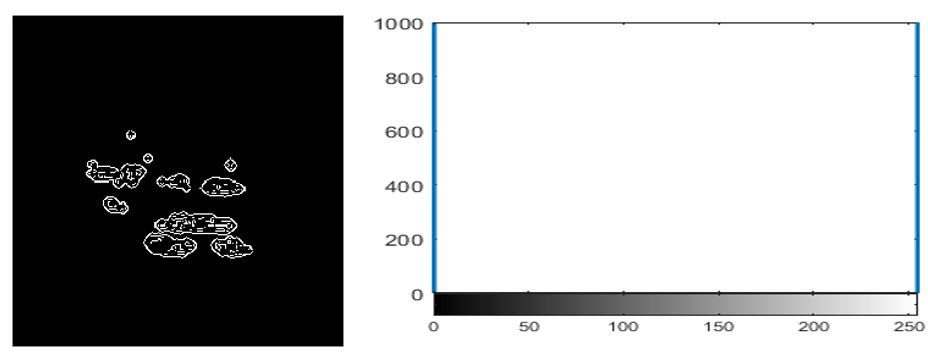}
         \caption{Canny Edge detection (Nevus)}
         \label{sd}
     \end{subfigure} 
     \\
     \begin{subfigure}[b]{0.48\textwidth}
         \centering
          \includegraphics[width=.80\linewidth]{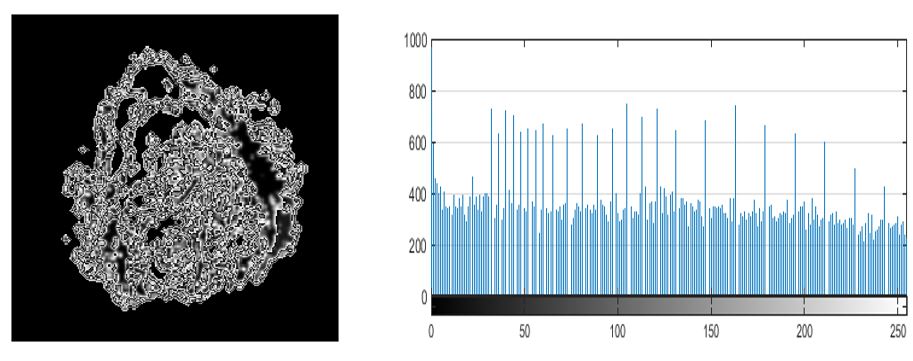}
         \caption{Unsigned Intensity (Melanoma)}
         \label{se}
     \end{subfigure}
     \hfill
     \begin{subfigure}[b]{0.48\textwidth}
         \centering
         \includegraphics[width=.80\linewidth]{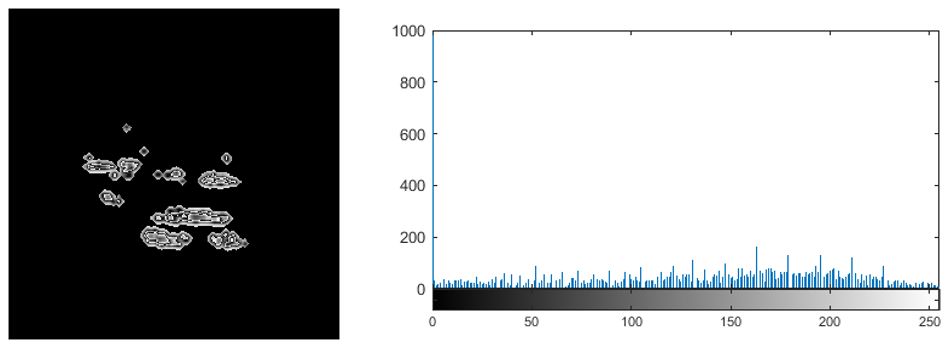}
         \caption{Unsigned Intensity (Nevus)}
         \label{sf}
     \end{subfigure}
     \\
         \begin{subfigure}[b]{0.48\textwidth}
         \centering
          \includegraphics[width=.80\linewidth]{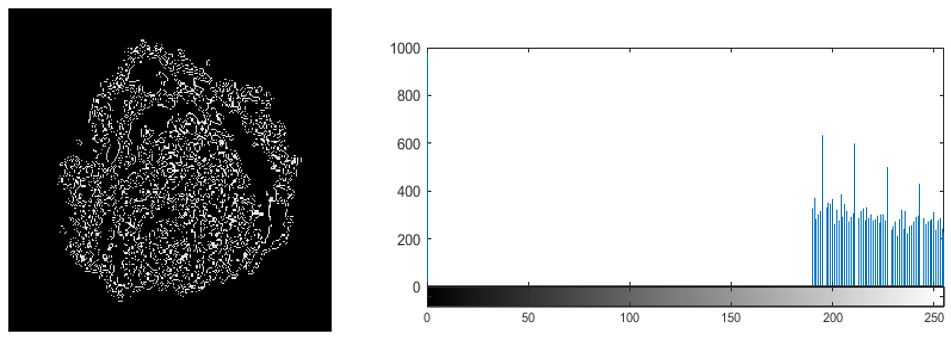}
         \caption{High Intense pixel Collection (Melanoma)}
         \label{sg}
     \end{subfigure}
     \hfill
     \begin{subfigure}[b]{0.48\textwidth}
         \centering
         \includegraphics[width=.80\linewidth]{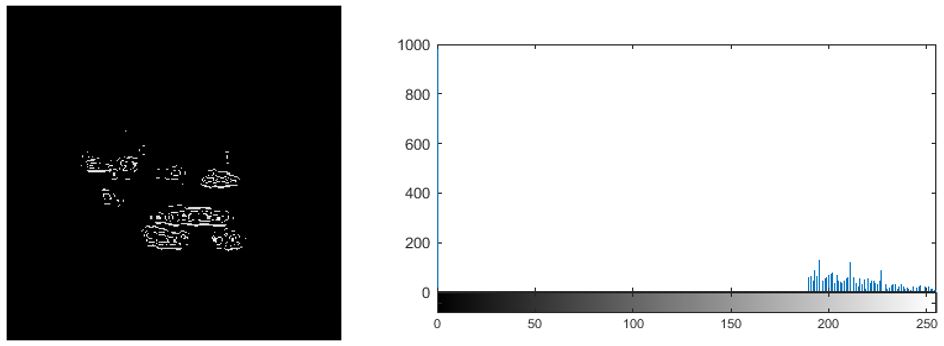}
         \caption{High Intense pixel Collection (Nevus)}
         \label{sh}
     \end{subfigure} 
  \caption{Melanoma and Nevus lesion Histogram comparison}
  \label{fig:pixelcomparison}
\end{figure*}

The intensity value estimation (IVE) model is one of the significant contributions of the work, as intensity levels act as one of the most distinctive features for object or region of interest identification \cite{43_khan2019classification}. In the IVE model, a multiplication between every pixel of the segmented lesion image and a constant value is performed using equation \ref{equ_image_const}. Here, $f$ is a 2D light intensity function, and $(x,y)$ denotes the spatial coordinates of the image. Then the normalization on the calculated image is used to distribute the intensity of the lesion region pixels using equation \ref{equ_image_norm}. The normalization process converts the gray-level intensity of the image by the range of 0 to 255 Fig. (\ref{se}) and Fig. (\ref{sf}). The histogram shows the pixel value between 0 and 255, whereas the edge detection of segmented lesion image histogram shows the pixel intensity in 0 or 255 Fig. (3c) and Fig. (3d). So, there are no intensity values between 0 and 255 in the edge detection image, which cuts off a few features from the image. Here, our proposed IVE model helps retain the lesion image's different intensity levels. These varieties of the intensity level of lesion images contain more information (color and texture) in the intensity level form than normal edge detection techniques. \\

Table \ref{table:5FoldValidationCanny} shows the experimental result of our proposed model using Canny Edge detection. Since the improvement of the effect is not significant, we resolved the model to distribute the gray-level intensity of the image using equation \ref{equ_image_norm}. It is observed that this modification can select the pixel values in the segmented lesion images that are greater or equal to a threshold value equation \ref{equ_image_IVE}. 
\begin{equation}
\label{equ_image_const}
    f(x,y) = f(x,y)\times \mathrm{Constant}, \ for \ every \ x, y
\end{equation}
\begin{equation}
\label{equ_image_norm}
    f(x,y) = normalize(f(x,y))
\end{equation}
\begin{equation}
\label{equ_image_IVE}
    f(x,y)=
    \begin{cases}
      f(x,y), & \text{if}\ f(x,y) > Threshold \\
      0, & \text{otherwise}
    \end{cases}
\end{equation}
Using the proposed high-intensity value estimation model produced more features in different intensity levels than the only edge detection of the segmented lesion Fig. (\ref{sg}) and Fig. (\ref{sh}). The model extracts the color and texture features in the form of the different intensity levels of the pigmentary lesion. In this aspect, choosing intense high pixels helps our model to differentiate Melanoma cells from Nevus cells. This decision experiment is depicted in Fig. \ref{fig:pixelcomparison}.

\subsection{Convolutional Neural Network}
\label{CNNLabel}
\begin{figure}[htb]
	\centering
		\includegraphics[height = 0.32\linewidth, width=.99\linewidth]{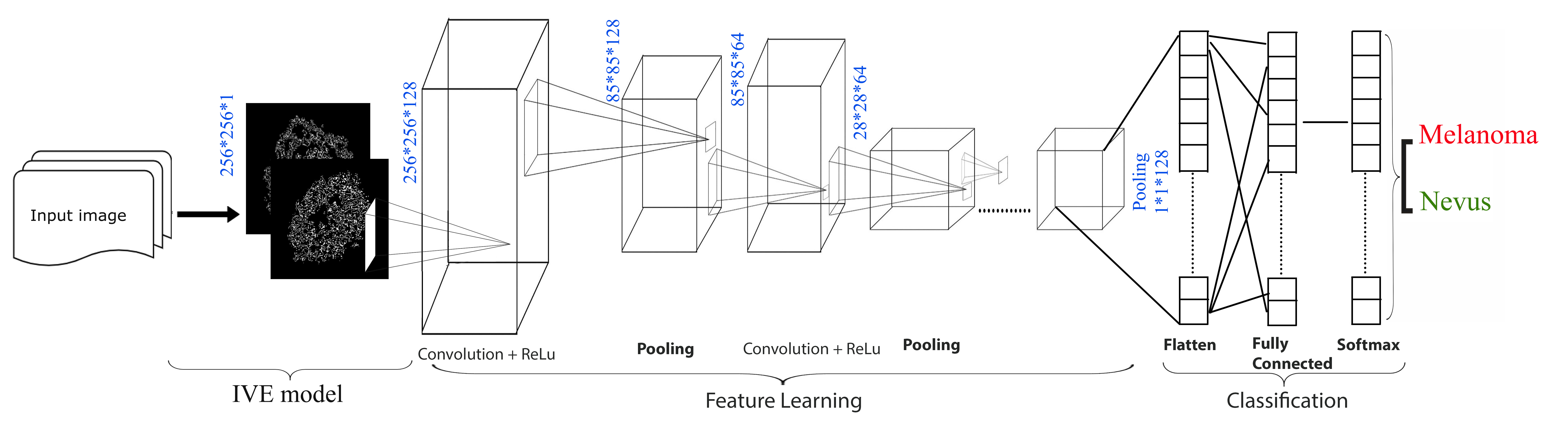}
	\caption{Proposed IVEwCNN model workflow}
	\label{FIG:CNNMODEL}
\end{figure}

\noindent It is proved that a simple artificial neural network (ANN) fails at a certain point, especially an over-fitting may arise due to the size of the image \cite{36_gogul2017flower}. So, CNN has tremendous advantages over ANN in image classification problems. CNN is a class of deep neural networks that makes a massive breakthrough in image classification, recognition, object detection, face recognition, and many more \cite{22_islam2020detection}. CNN helps to auto detect important features and extract features from the images, which may be helpful in the image classification task \cite{12_refianti2019classification}. It extracts features from images using filters and reduces the number of learnable parameters using the pooling technique. The CNN model comprises five stages of neural layers in its structure: the input layers, convolutional layer (Convo + ReLu), pooling layer, fully connected layer, and output layer. The proposed CNN model workflow is depicted in Fig. \ref{FIG:CNNMODEL}. 

CNN model accepts the input as a 3-dimensional matrix ($Width \times Height \times Dimension$) to the input layer. Dimension defines the number of channels contains in that image. Dimension for a grayscale image is 1, and for RGB, it is 3. The first layer of the CNN model is the input layer, where the input image with size $256 \times 256 \times 1$ is given. Then input image is passed to the convolutional layer, where filters are applied to the original images. The filters are slid over the receptive fields of the same input image by a stride and continue through the whole image. The convolutional layer uses the ReLu activation function to zero all negative values. A pooling layer is used after each convolutional layer to reduce the spatial volume of the input image. The fully connected layer takes the output from the previous layer and flattens them to convert inputs into a single vector. This layer involves weights, biases, neurons, and activation functions and is responsible for updating weights in the training session. Here, we used the ReLu activation function to produce output from each layer. The output layer is the final layer that uses the Softmax activation function to estimate Melanoma skin cancer and Nevus mole probability.

\section{Dataset}
\label{datasetLabel}
\begin{figure}[!htb]
     \centering
     \begin{subfigure}[b]{0.23\textwidth}
         \centering
          \includegraphics[height = 0.85\linewidth, width=.95\linewidth]{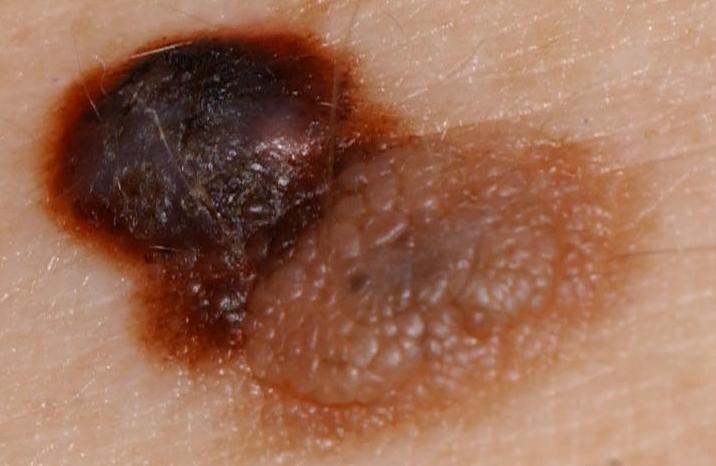}
          \caption{}
     \end{subfigure}
     \begin{subfigure}[b]{0.23\textwidth}
         \centering
         \includegraphics[height = 0.85\linewidth,width=.95\linewidth]{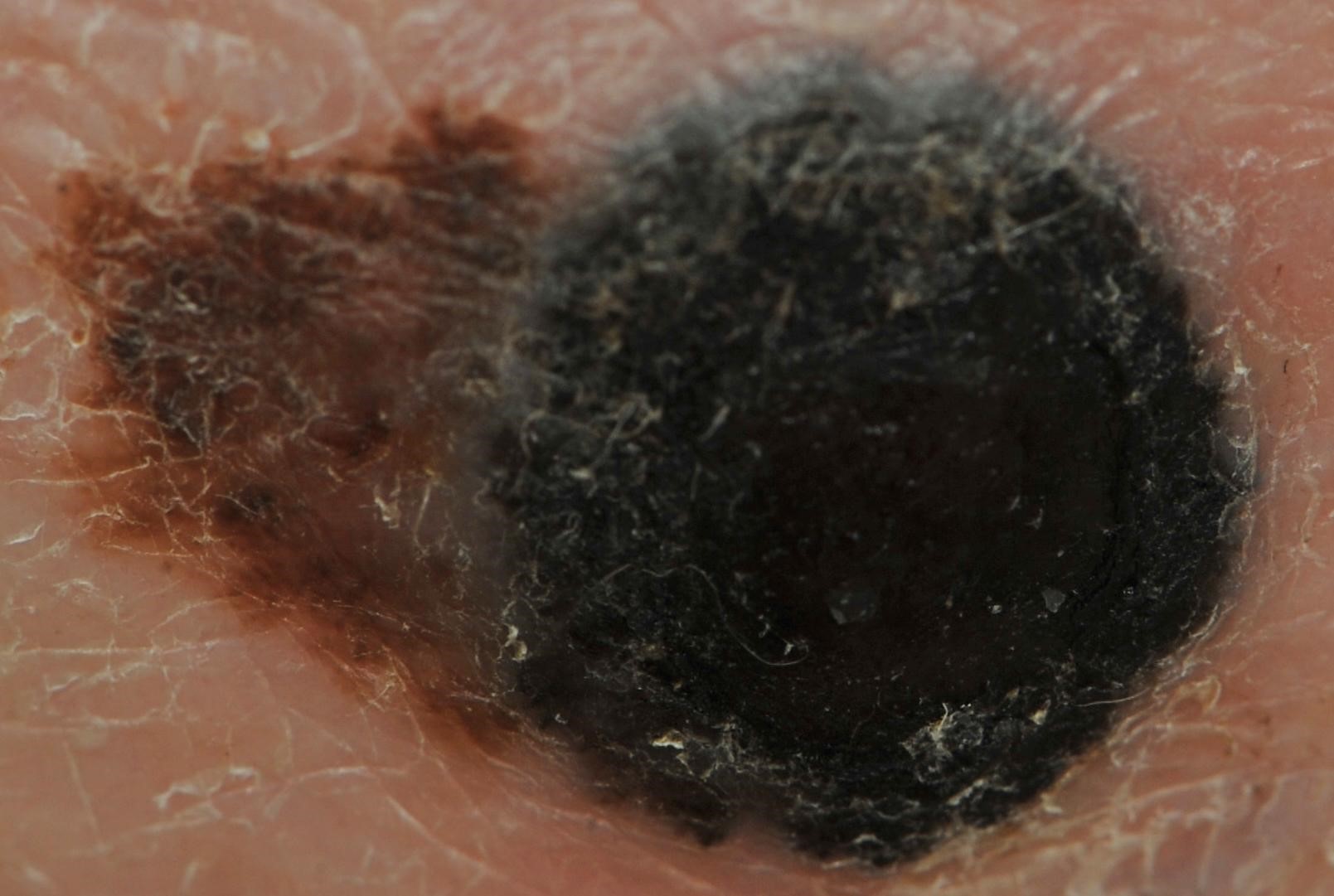}
         \caption{}
     \end{subfigure}
     \begin{subfigure}[b]{0.23\textwidth}
         \centering
          \includegraphics[height = 0.85\linewidth, width=.95\linewidth]{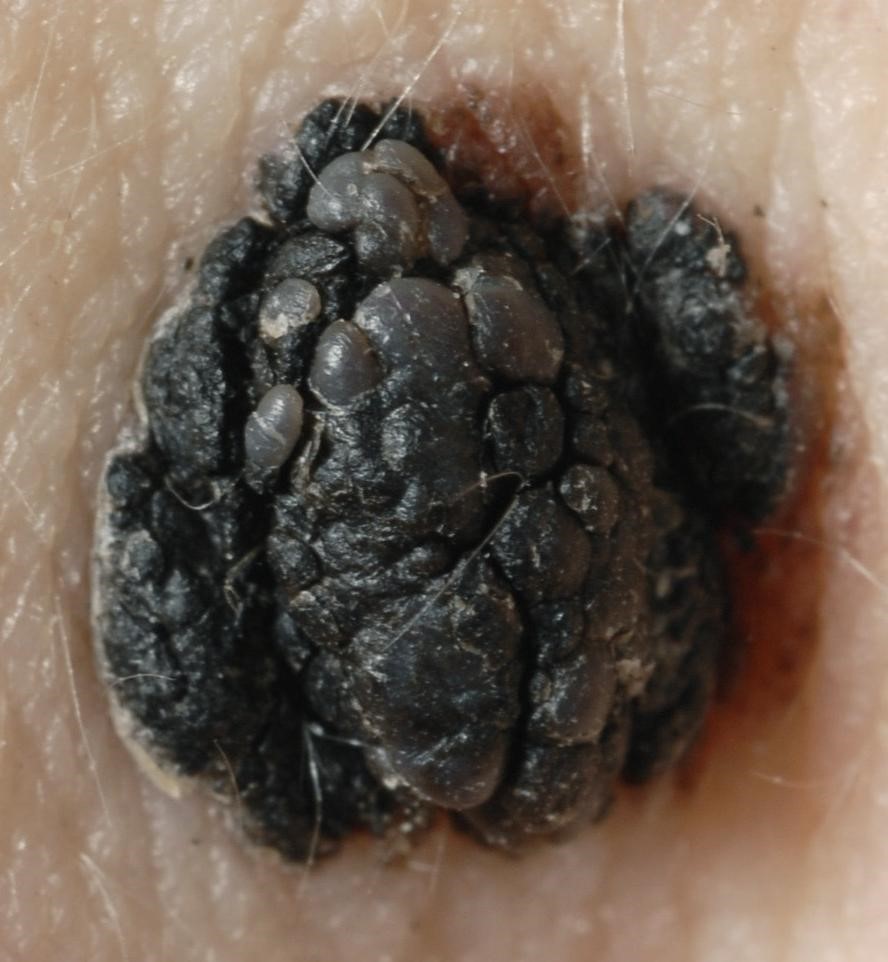}
          \caption{}
     \end{subfigure}
     \begin{subfigure}[b]{0.23\textwidth}
         \centering
         \includegraphics[height = 0.85\linewidth,width=.95\linewidth]{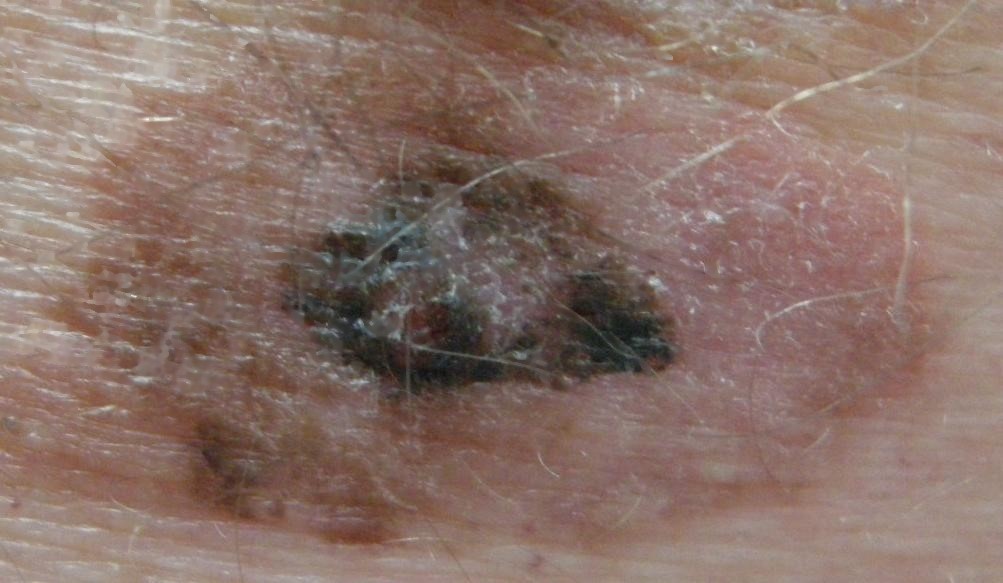}
         \caption{}
     \end{subfigure} 
  \caption{Melanoma Skin-Cancer images}
  \label{melanomaDatasetLabel}
\end{figure}

\begin{figure}[!htb]
     \centering
     \begin{subfigure}[b]{0.23\textwidth}
         \centering
          \includegraphics[height = 0.80\linewidth, width=.95\linewidth]{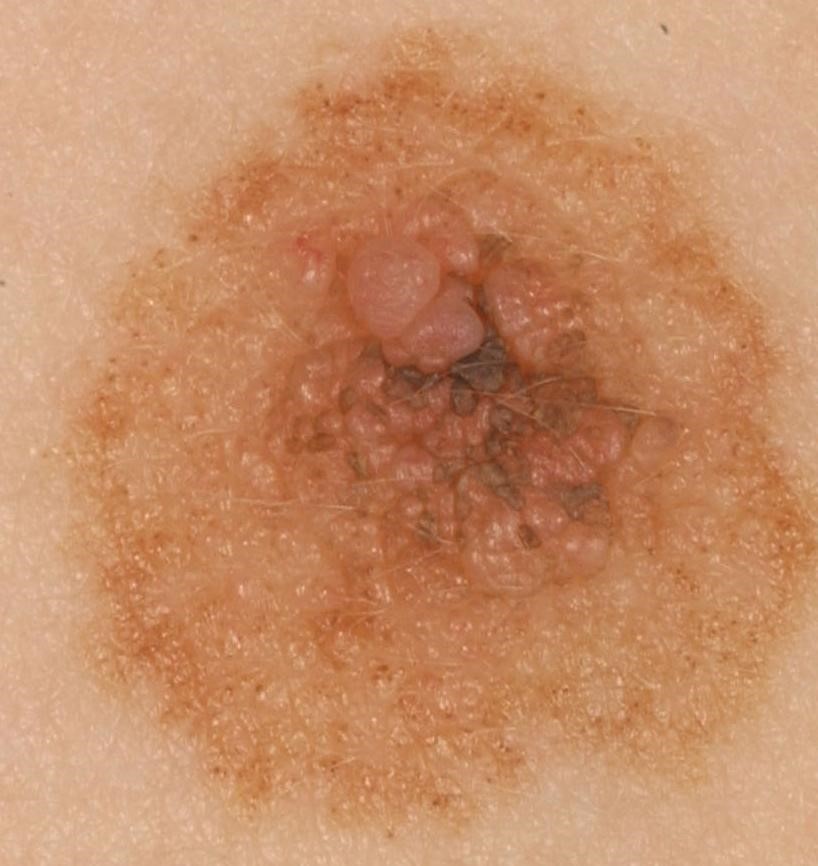}
          \caption{}
     \end{subfigure}
     \begin{subfigure}[b]{0.23\textwidth}
         \centering
         \includegraphics[height = 0.80\linewidth,width=.95\linewidth]{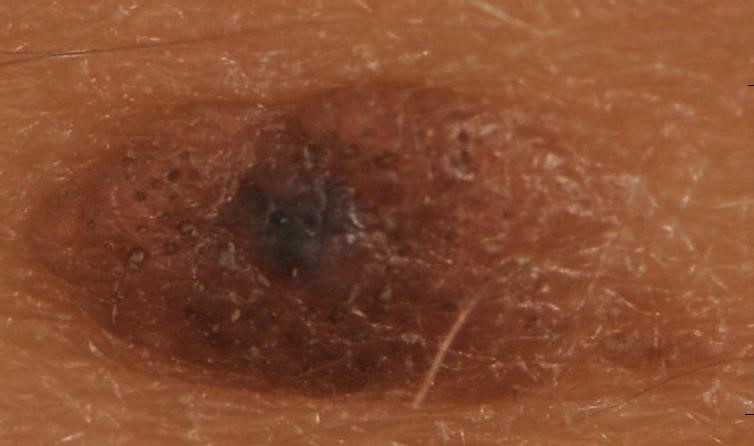}
         \caption{}
     \end{subfigure}
     \begin{subfigure}[b]{0.23\textwidth}
         \centering
          \includegraphics[height = 0.80\linewidth, width=.95\linewidth]{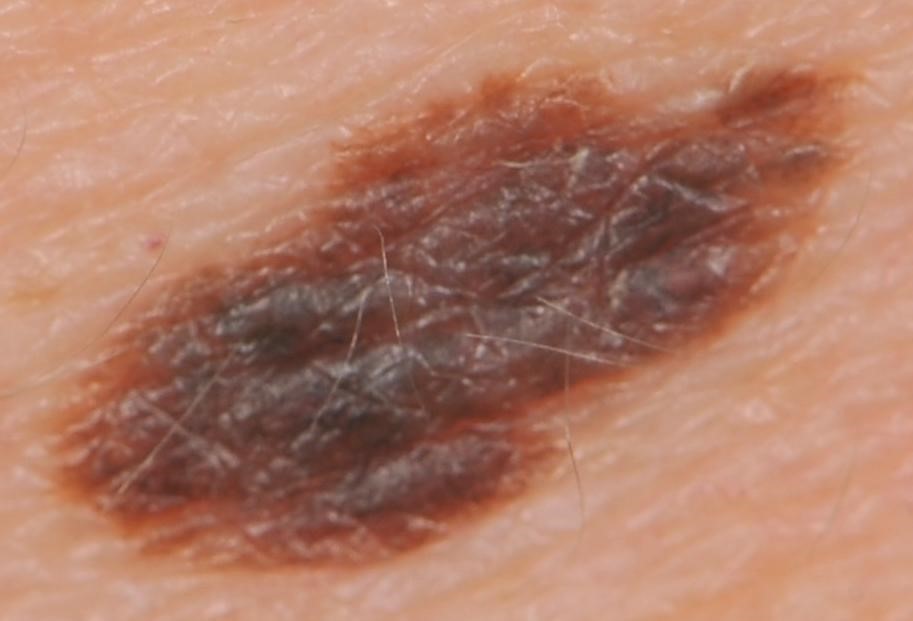}
          \caption{}
     \end{subfigure}
     \begin{subfigure}[b]{0.23\textwidth}
         \centering
         \includegraphics[height = 0.80\linewidth,width=.95\linewidth]{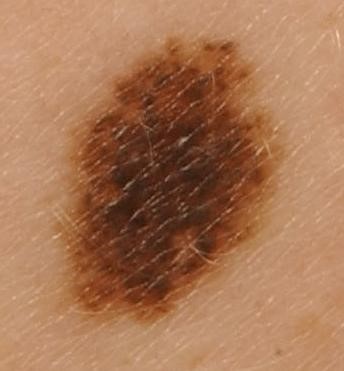}
         \caption{}
     \end{subfigure} 
  \caption{Nevus Mole Images}
  \label{nevusDatasetLabel}
\end{figure}

The proposed system is evaluated on a publicly available standard MED-NODE melanoma dataset, which contains high-resolution skin lesion images \cite{4_giotis2015med}. MED-NODE dataset is the subset of digital image archives of 50,000 images collected by the Department of Dermatology of the University Medical Center Groningen (UMCG). All of these pictures have been examined and assessed by a dermatologist as being of the highest caliber \cite{4_giotis2015med}. These photographs were captured in JPEG format with a Nikkor lens on a Nikon D3 and Nikon D1x camera from a distance of around 33 cm from the skin lesion area. Fig. \ref{melanomaDatasetLabel} and \ref{nevusDatasetLabel} display some of the images from the MED-NODE dataset.\\ 
The soundness of the dataset has been ensured by considering the following criteria \cite{4_giotis2015med}.
\begin{itemize}
    \item MED-NODE dataset (170 images) is created by randomly chosen images and the relevant patient instances could not possibly be identified.
    \item Superficial spreading melanoma and nevi are included in the dataset.
    \item Each picture originated from a different patient (Aside from the image, which shows how the disease significantly differs in various bodily areas).
    \item Each picture is sharp and properly exposed to annotate correctly.
    \item Each picture has represented the group it belongs to.
\end{itemize}

\section{Experimental Analysis}
\label{expAnalysis}
Some of the dataset image resolutions are above ($N_r \times N_c$) (where $N_r = 256$ and $N_c = 256$), which needs a high cost of computing power. Since the rescaled lesion images would be better for deep neural networks \cite{47_salih2020skin}, 
we used pre-processing steps to resize the images into ($N_r \times N_c$) and remove unwanted artifacts. As the low amount of data could mislead and overfit the deep neural network model, the augmentation technique is applied to increase the amount of data and the diversity of the data \cite{41_shorten2019survey} to reduce the data misleading and overfitting. Then the IVE model is applied to acquire high intense pixel values from every segmented lesion image, which allows for keeping important information about Melanoma and Nevus mole features. Feeding the significant feature value to CNN achieves a better result.\\ 

Authors \cite{review_a_corr_3_guyon1997scaling} discovered that for the training/validation set, the number of available adjustable parameters for the dataset should be inversely proportional to the square of the significant portion of patterns. As a result, the 80/20 split was chosen to avoid overtraining the deep neural network. Most of the state-of-the-art methods \cite{46_jianu2019automatic,7_mukherjee2019malignant, 4_giotis2015med,3_nasr2016melanoma} that use the same or smaller dataset used 80\% data for training and 20\% data for testing purposes. In the experiment, the data augmentation technique is used to enhance the dataset. After the data augmentation, 80\% images are employed to train the algorithm, and the rest 20\% are used for validation and testing purposes. To avoid biased performance results we chose average results from the 5-fold cross-validation on our proposed model.


\subsection{Training and Testing}
At the beginning of our system, it starts to train our model with 80\% train images, and the rest 20\% is used for testing purposes. Our system split the dataset such that no overlapping occurs in the train and test data set. We set 40 as the number of epochs to train our model. For optimization purposes, we used Adam Optimization and Sparse Categorical Cross entropy loss to calculate the loss of the CNN model. We implemented our model in the google cloud platform using Colab Notebook, which provides a 12 GB NVIDIA Tesla K80 GPU that can be used for up to 12 hours continuously and is also highly integrated with Google Drive. It also offered TPU recently for free. The full CNN model is depicted in Table \ref{table:CNNARCH}. The proposed CNN architecture is comparable with the other well-known DNN models that are implemented for skin cancer detection. To achieve the best experimental results DNN model must have the minimum number of these proposed layers with the IVE model in the processed image.
\begin{table*}[htb]
\centering
\caption{Architecture of CNN Layers.}\label{table:CNNARCH}
\begin{tabular}{lllr}
\hline\noalign{\smallskip}
Layers & Output Size & Kernel Size & Activation Function\\
\noalign{\smallskip}\hline\noalign{\smallskip}
Input Layer & 256×256×1 & - & - \\
Conv2D & 256×256×128 & 7×7 & ReLu Activation \\
MaxPooling2D & 85×85×128 & 3×3 (strides = 3) & - \\
Conv2D & 85×85×64 & 4×4 & ReLu Activation \\
MaxPooling2D & 28×28×64 & 3×3 (strides = 3) & - \\

Conv2D	& 28×28×32	& 5×5	& ReLu Activation\\
MaxPooling2D	& 14×14×32	& 2×2 (strides = 2)& 	-\\
Conv2D	& 14×14×128	& 6×6	& ReLu Activation\\
MaxPooling2D	& 4×4×128	& 3×3 (strides = 3)	& -\\
Conv2D	& 4×4×32	& 5×5	& ReLu Activation\\
MaxPooling2D	& 2×2×32	& 2×2 (strides = 2)	& -\\
Conv2D	& 2×2×128	& 6×6	& ReLu Activation\\
MaxPooling2D	& 1×1×128	& 2×2 (strides = 2)	& -\\
Flatten	& 128	& -	& -\\
Dense182	& 512	& -	& ReLu Activation\\
Dense183	& 128	& -	& ReLu Activation\\
Dense184	& 64	& -	& ReLu Activation\\
Dense185	& 512	& -	& ReLu Activation\\
Dense186	& 512	& -	& ReLu Activation\\
Dense187	& 64	& -	& ReLu Activation\\
Dense188	& 2	& -	& Softmax Activation\\
\noalign{\smallskip}\hline
\end{tabular}
\end{table*}

\subsection{Performance Evaluation}
\label{secExp}
The proposed IVE model was used to estimate the high-intensity pixel values from the training dataset, which are incorporated into our developed CNN model to detect and classify melanoma skin cancer and nevus mole. Table \ref{table:5FoldValidationCanny} shows the 5-fold cross-validation experimental results of our proposed model using Canny Edge detection and Table \ref{table:5FoldValidation} shows the 5-fold cross-validation experimental results of our proposed model using intensity value estimation with CNN (IVEwCNN). It is demonstrated that pixel intensity levels serve as one of the distinguishing features for identifying an object or region of interest, particularly when considering skin cancer. Textural features are regularly used in image classification because they enhance the classification of nevus and melanoma by computing the irregularity of their structure \cite{43_khan2019classification}. Since choosing the high-intense pixels gives more information, the model can easily differentiate Melanoma cells from Nevus cells more preciously. Therefore the proposed model shows a better result when compared with others. \\

\begin{table*}[ht]
\caption{5-Fold cross-validation experimental result using Canny Edge detection with CNN}\label{table:5FoldValidationCanny}
\centering
\begin{tabular}{llllllr}
\hline\noalign{\smallskip}
Metrics	& Fold-1	& Fold-2	& Fold-3	& Fold-4	& Fold-5	& Average\\
\noalign{\smallskip}\hline\noalign{\smallskip}
Sensitivity	& 0.50 & 0.75 & 0.375 & 0.438 & 0.688 & 0.55\\
Specificity	& 0.789 & 0.684 & 0.842 & 0.737 & 0.842 & 0.779\\
PPV	& 0.667 & 0.667 & 0.667 & 0.583 & 0.786 & 0.674\\
NPV	& 0.652 & 0.765 & 0.615 & 0.609 & 0.762 & 0.680\\
Accuracy	& 0.657 & 0.714 & 0.629 & 0.60 & 0.771 & 0.674\\
\noalign{\smallskip}\hline
\end{tabular}
\end{table*}

\begin{table*}[!ht]
\centering
\caption{5-Fold cross-validation experimental result using Intensity Value Estimation with CNN (IVEwCNN)}\label{table:5FoldValidation}
\centering
\begin{tabular}{llllllr}
\hline\noalign{\smallskip}
Metrics	& Fold-1	& Fold-2	& Fold-3	& Fold-4	& Fold-5	& Average\\
\noalign{\smallskip}\hline\noalign{\smallskip}
Sensitivity	& 1.00	& 1.00	& 1.00	& 0.875	& 0.813	& 0.9376\\
Specificity	& 0.894	& 1.00	& 1.00	& 0.789	& 0.895	& 0.9156\\
PPV	& 0.889	& 1.00	& 1.00	& 0.778	& 0.867	& 0.9068\\
NPV	& 1.00	& 1.00	& 1.00	& 0.894	& 0.850	& 0.9488\\
Accuracy	& 0.943	& 1.00	& 1.00	& 0.829	& 0.857	& 0.9258\\
\noalign{\smallskip}\hline
\end{tabular}
\end{table*}

\begin{table*}[ht]
\centering
\caption{Experimental result evaluation of the Proposed Methodology with the State-of-the-art Methods }\label{table:BESTresultLabel}
\centering
\begin{tabular}{l p{1.3cm} p{1cm} p{1.1cm} p{.6cm} c}
\hline
\toprule
    \multirow{2}{*}{\bfseries Methods} & 
    \multicolumn{5}{c}{\bfseries Metrics}\\ 
    \cmidrule(lr){2-6}
 & \parbox[t]{1.0cm}{\centering Sensitivity \\ (Recall)} & Specificity &  \parbox[t]{1.5cm}{\centering PPV\\(Precision)} & NPV & Accuracy \\ 
    \cmidrule(lr){1-6}
    Texture descriptor \cite{4_giotis2015med}	& 0.62	& 0.85	& 0.74	& 0.77	& 0.76\\
    Color descriptor \cite{4_giotis2015med}	& 0.74	& 0.72	& 0.64	& 0.81	& 0.73\\
    \pbox{20cm}{Illumination Correction \cite{3_nasr2016melanoma}}	& 0.81	& 0.80	& 0.75	& 0.86	& 0.81\\
   \pbox{20cm}{Optimized NN \\using PSO \cite{7_mukherjee2019malignant}}	& 0.86	& 0.86	& -	& -	& 0.86\\
   \pbox{20cm}{S. R. S. Jianu et al. \cite{46_jianu2019automatic}}	& 0.72	& 0.89	& 0.87	& 0.76	& 0.81\\
   \pbox{20cm}{S. Mukherjee et al. \\(raw feature only) \cite{45_mukherjee2020malignant}}	& 0.87	& 0.73	& -	& -	& 0.83\\
   \pbox{20cm}{S. Mukherjee et al. \\(PCA feature only) \cite{45_mukherjee2020malignant}}	& 0.87	& 0.87	& -	& -	& 0.87\\
    \pbox{20cm}{Proposed Methodology \\ (Canny Edge detection)}	& \textbf{0.65}	& \textbf{0.78}	& \textbf{0.67}	& \textbf{0.68}	& \textbf{0.67}\\
    Proposed Methodology\\ (IVEwCNN)	& \textbf{0.94}	& \textbf{0.92}	& \textbf{0.91}	& \textbf{0.95}	& \textbf{0.93}\\
\noalign{\smallskip}
\bottomrule
\end{tabular}
\end{table*}

From Table \ref{table:5FoldValidationCanny} and Table \ref{table:5FoldValidation}, we can see that our proposed (IVEwCNN) model performs significantly better in terms of every described performance evaluation metric than the proposed model with Canny Edge detection. We used the same threshold value for all the performance evaluation metrics in this experiment. For experimental evaluation of the algorithm, the proposed model was compared with some existing works described in \cite{4_giotis2015med,3_nasr2016melanoma,7_mukherjee2019malignant,46_jianu2019automatic,45_mukherjee2020malignant} as they all worked with the same or smaller dataset and evaluated their system performance. For performance measurement, the proposed model was evaluated on five commonly employed metrics (sensitivity, specificity, PPV, NPV, and accuracy) that are widely used in classification problems. The performance evaluation metrics considered in this work can be defined as below:
\begin{eqnarray}
\label{SensEquation2}
Sensitivity = \frac{true~detected ~melanoma ~cases}{all~ melanoma ~cases}
\end{eqnarray}
\begin{eqnarray}\label{SpeEquation3}
Specificity = \frac{true ~detected~non~melanoma~cases}{all~non~melanoma~ cases}
\end{eqnarray}
\begin{eqnarray}\label{PPVequation4}
PPV = \frac{true ~detected ~melanoma ~cases}{detected ~melanoma ~cases}
\end{eqnarray}
\begin{eqnarray}\label{NPVequation5}
NPV = \frac{true ~detected ~non~melanoma ~cases }{detected ~non~melanoma ~cases}
\end{eqnarray}
\begin{eqnarray}\label{AccuEquation6}
Accuracy = \frac{true ~detected ~cases}{all ~cases}
\end{eqnarray}
The proposed IVE model performance and the Canny Edge detection technique performances are compared with other state-of-the-art methods that were evaluated on the same or smaller dataset described in Table \ref{table:BESTresultLabel}. The values in Table \ref{table:BESTresultLabel} are updated to two decimal points to compare with other state-of-the-art methods. In Table \ref{table:BESTresultLabel} the results in bold format show the experimental results. Our proposed methodology (IVEwCNN) shows better performance evaluation metrics. We can conclude that the effective use of the CNN model with a well-processed image generates a superior result when we use high intense pixel values from the segmented lesion skin image.
\section{Discussion}
Instead of conventional visual observations, an efficient expert system is developed to assist expert physicians in the early-stage detection and classification of Melanoma skin cancer. The proposed IVEwCNN model adopts the new intensity value estimation (IVE) technique in which the high intense pixel values are calculated from each segmented lesion image after rescaling the image. Most of the work is done by conducting lesion segmentation and detecting the edges of the segmented lesion, which is then fed into a deep neural network. This contrast enhancement and texture analysis increase the discrimination between the intensity of values of an image that improves the overall performance \cite{24_Zheng2008}. Our proposed method performs this more precisely and accurately using three different stages. Firstly, in preprocessing step, a new image is generated consisting of a shape of ($N_r \times N_c$) where the lower dimension of segmented image pixels are mapped to retain the original lesion shape. Keeping the lesion shape the same helps retain information about the lesion area, size, and border.

\begin{figure}[!htb]
     \centering
     \begin{subfigure}[b]{0.23\textwidth}
         \centering
          \includegraphics[height = 0.85\linewidth, width=.95\linewidth]{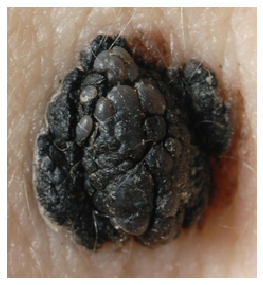}
          \caption{}
     \end{subfigure}
     \begin{subfigure}[b]{0.23\textwidth}
         \centering
         \includegraphics[height = 0.85\linewidth,width=.95\linewidth]{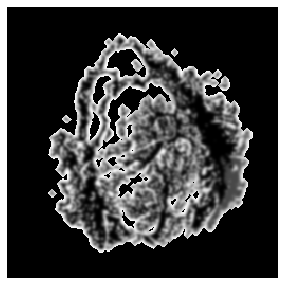}
         \caption{}
     \end{subfigure}
     \begin{subfigure}[b]{0.23\textwidth}
         \centering
          \includegraphics[height = 0.85\linewidth, width=.95\linewidth]{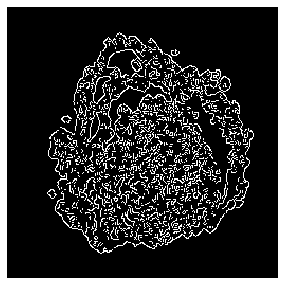}
          \caption{}
     \end{subfigure}
     \begin{subfigure}[b]{0.23\textwidth}
         \centering
         \includegraphics[height = 0.85\linewidth,width=.95\linewidth]{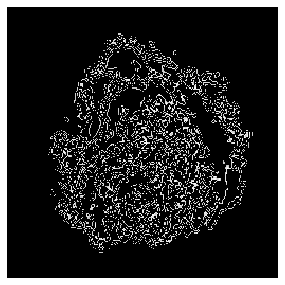}
         \caption{}
     \end{subfigure} 
  \caption{Melanoma Skin-Cancer images (a) Melanoma skin lesion (b) Resized segmented lesion (c) Canny Edge detection (d) IVE model output}
  \label{FIG:melanomaDiscussLabel}
\end{figure}

\begin{figure}[!htb]
     \centering
     \begin{subfigure}[b]{0.23\textwidth}
         \centering
          \includegraphics[height = 0.80\linewidth, width=.95\linewidth]{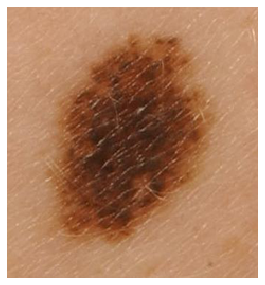}
          \caption{}
     \end{subfigure}
     \begin{subfigure}[b]{0.23\textwidth}
         \centering
         \includegraphics[height = 0.80\linewidth,width=.95\linewidth]{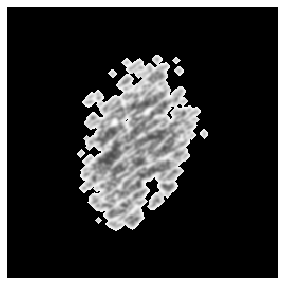}
         \caption{}
     \end{subfigure}
     \begin{subfigure}[b]{0.23\textwidth}
         \centering
          \includegraphics[height = 0.80\linewidth, width=.95\linewidth]{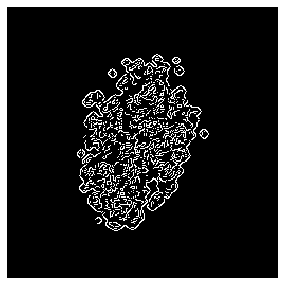}
          \caption{}
     \end{subfigure}
     \begin{subfigure}[b]{0.23\textwidth}
         \centering
         \includegraphics[height = 0.80\linewidth,width=.95\linewidth]{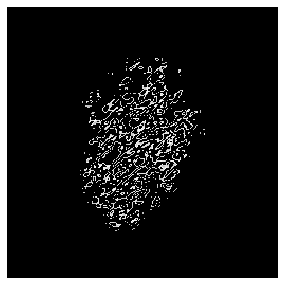}
         \caption{}
     \end{subfigure} 
  \caption{Nevus Mole Images (a) Nevus skin lesion (b) Resized segmented lesion (c) Canny Edge detection (d) IVE model output}
  \label{FIG:nevusDiscussLabel}
\end{figure}

Then the IVE model is applied as shown in Fig. \ref{FIG:melanomaDiscussLabel} and Fig. \ref{FIG:nevusDiscussLabel} to compute the high-intensity pixels that store descriptive features on object or region of interest identification \cite{43_khan2019classification}.
The proposed model (IVEwCNN) shows a significant difference between the outcome of Canny Edge detection and existing works. Table \ref{table:5FoldValidationCanny} depicts the best outcome of Canny Edge detection with CNN is 0.75 for sensitivity at Fold-2, 0.842 for specificity at Fold-3 and Fold-5, 0.786 for PPV at Fold-5, 0.765 for NPV at Fold-2, and 0.771 for accuracy at Fold-5. The result of existing works is shown in Table \ref{table:BESTresultLabel} where the best result was achieved 0.8744 for sensitivity \cite{45_mukherjee2020malignant}, 0.89 for specificity \cite{46_jianu2019automatic}, 0.8674 for PPV \cite{46_jianu2019automatic}, 0.86 for NPV \cite{3_nasr2016melanoma} and 0.8718 for Accuracy \cite{45_mukherjee2020malignant}. Whereas our proposed model exhibits the average outcome of IVEwCNN is 0.936 for sensitivity, 0.9156 for specificity, 0.9068 for PPV, 0.9488 for NPV, and 0.9258 for accuracy, as shown in Table \ref{table:5FoldValidation} which outperforms the conventional edge detection based techniques.

\section{Conclusion}
\label{secConc}
This paper presented intensity value estimation with a convolutional neural network (IVEwCNN) based algorithm for detection and classification. The method led the system to achieve high accuracy, sensitivity, specificity, precision, and NPV to detect and classify melanoma skin cancer and nevus mole. As well, pre-processed images increase the learnability of any system. Here, we chose to take the pixels with higher intensity than a threshold value from the segmented lesion image instead of edge detection. The technique preserves more features than only edge detection. So, this helps our system to increase learnability; hence this predicts melanoma, skin cancer, and nevus mole more accurately. Our proposed system takes 39 (average) seconds to detect and predict melanoma skin cancer and nevus mole. Finally, to evaluate the proposed system performance, we considered the five most popular performance evaluation metrics and compared them with some notable existing works on the same or smaller dataset. Due to machine limitations, large datasets were not considered in this experiment to evaluate the proposed IVE model. Hence, the proposed model's performance is only evaluated by comparing it to the state-of-the-art models that used the same or smaller dataset for their model. The experimental comparison exhibits that the proposed algorithm shows better results than all the state-of-the-art models. The proposed automatic deep learning system can be implemented in dermatological diagnosis to aid doctors in early melanoma detection. Before going for a clinical application, our proposed system needs to train the model with an improved large dataset. Future work could include feature selection, feature dimension reduction, and optimizing the CNN model to increase efficiency and decrease model training time. 

\section*{Acknowledgments}
The authors thank the Department of Computer Science and Engineering of Dhaka University of Engineering \& Technology, Gazipur, for providing research support to continue the research work.

\bibliographystyle{cs-agh}
\bibliography{bibliography}

\end{document}